# Pituitary Adenoma Volumetry with 3D Slicer

**Jan Egger[1,2,3]\*[9], Tina Kapur[1][9], Christopher Nimsky[2], Ron Kikinis[1]**

**1** Department of Radiology, Brigham and Women's Hospital, Harvard Medical School, Boston, Massachusetts, United States of America, **2** Department of Neurosurgery, University Hospital of Marburg, Marburg, Germany, **3** Department of Mathematics and Computer Science, The Philipps-University of Marburg, Marburg, Germany

## Abstract

In this study, we present pituitary adenoma volumetry using the free and open source medical image computing platform for biomedical research: (3D) Slicer. Volumetric changes in cerebral pathologies like pituitary adenomas are a critical factor in treatment decisions by physicians and in general the volume is acquired manually. Therefore, manual slice-by-slice segmentations in magnetic resonance imaging (MRI) data, which have been obtained at regular intervals, are performed. In contrast to this manual time consuming slice-by-slice segmentation process Slicer is an alternative which can be significantly faster and less user intensive. In this contribution, we compare pure manual segmentations of ten pituitary adenomas with semi-automatic segmentations under Slicer. Thus, physicians drew the boundaries completely manually on a slice-by-slice basis and performed a Slicer-enhanced segmentation using the competitive region-growing based module of Slicer named GrowCut. Results showed that the time and user effort required for GrowCut-based segmentations were on average about thirty percent less than the pure manual segmentations. Furthermore, we calculated the Dice Similarity Coefficient (DSC) between the manual and the Slicer-based segmentations to proof that the two are comparable yielding an average DSC of $81.97\pm3.39\%$.

**Citation:** Egger J, Kapur T, Nimsky C, Kikinis R (2012) Pituitary Adenoma Volumetry with 3D Slicer. PLoS ONE 7(12): e51788. doi:10.1371/journal.pone.0051788

**Editor:** Arrate Muñoz-Barrutia, University of Navarra, Spain

**Received** September 20, 2012; **Accepted** November 8, 2012; **Published** December 11, 2012

**Copyright:** © 2012 Egger et al. This is an open-access article distributed under the terms of the Creative Commons Attribution License, which permits unrestricted use, distribution, and reproduction in any medium, provided the original author and source are credited.

**Funding:** This work was supported by National Institutes of Health (NIH) grant P41EB015898. Its contents are solely the responsibility of the authors and do not necessarily represent the official views of the NIH. The funders had no role in study design, data collection and analysis, decision to publish, or preparation of the manuscript.

**Competing Interests:** The authors have declared that no competing interests exist.

\* E-mail: egger@bwh.harvard.edu

[9] These authors contributed equally to this work.

## Introduction

Tumors of the sellar region – primarily pituitary adenomas – represent 10% to 25% of all intracranial neoplasms and adenomas comprising the largest portion with an estimated prevalence of approximately 17% [1] and [2]. Adenomas can be classified according to several criteria including the size or the hormone secretion, like secreted hormones include cortisol (Cushing's disease) and treatment is in general followed by a decrease of prolactine levels and tumor volume, whereas the first choice of treatment for Cushing's disease remains surgery [3] and [4]. However, for hormone-inactive mircroadenomas, which are less than 1 cm in diameter, there is no need for a direct surgical resection and the follow-up examinations contain endocrine and ophthalmological evaluation, and magnetic resonance imaging mainly performed in one year intervals. In contrast to a wait-and-scan strategy which is no longer indicated, the microsurgical removal becomes the treatment of choice, in the case of continuous tumor volume progress, which has to be evaluated each time. Therefore, image analysis that includes segmentation and registration of these successive scans is useful in the accurate measurement of tumor progression.

In this section, related work in the field of supporting pituitary adenoma surgery is summarized. Other authors working in this field are Neubauer et al. [5] and [6] and Wolfsberger et al. [7] who investigated a virtual endoscopy system called STEPS. STEPS is designed to aid surgeons performing pituitary surgery, thereby using a semi-automatic segmentation approach which is based on the so-called watershed-from-markers method. This segmentation method technique uses manually defined markers in the object of interest – in this case the pituitary adenoma – and the background. The watershed-from-markers method is very computationally intensive, but Felkel et al. [8] introduced a memory efficient and fast implementation which can also be extended to 3D. Zukic et al. [9] developed a deformable model based approach that uses balloon inflation forces [10] for the segmentation of pituitary adenomas. The balloon inflation forces are used to expand a mesh iteratively incorporating different features for the vertex movement calculation: Vertices with lower curvature are moved outwards by a larger amount, thus stimulating smoother meshes. Vertices with high angle between normal and center vertex- vector are inflated by a smaller amount, in order to penalize protrusions. A recently introduced graph-based method for pituitary adenoma segmentation starts by setting up a directed and weighted 3D graph from a user-defined seed point that is located inside the pituitary adenoma [11]. Accordingly graph construction, the minimal cost closed set on the graph is computed via a polynomial time s-t cut [12]. The graph-based approach samples along rays that are sent through the surface points of a polyhedron [13] to generate the graph (note: the center of the polyhedral user-defined seed point that is located inside the pituitary adenoma). A novel multi-scale sheet enhancement measure that has been applied to paranasal sinus bone segmentation has been presented by Descoteaux et al. [14]. For the simulation of pituitary surgery, this measure has essential properties, which should be incorporated in the computation of anatomical models. However, if the volume of pituitary adenomas is analyzed over a long time of period for clinical studies, this is in general done via manual slice-by-slice segmentation, or sometimes semi-automatically supported by a software tool. Then, the three-dimensional tumor volume is calculated out of the single 2D contours, the amount of slices and the slice thickness [15], [16] and [17]. The growth of on-functioning pituitary adenomas in patients referred for surgery for example, has been studied by Honegger et al. [18], by calculating the three-dimensional tumor volume from the two-dimensional contours that have been manually outlined on each slice. Pituitary adenoma volume changes after gamma knife radiosurgery (GKRS) have been studied by Pamir et al. [19]. Therefore, the magnetic resonance imaging (MRI)-based volumetric analysis of the pituitary adenomas was done by using GammaPlan software from Elekta Instruments (Atlanta, GA) for tumor volume at the time of treatment. For the tumor volume on the follow-up MRI scans software from Imaging Inc. (Waterloo, Canada) was used. However, no further details have been provided how time-consuming and precise this procedure is by comparing it with ground truth segmentations from experts, e.g. manual slice-by-slice segmentations. Jones and Keogh [20] introduced a simple technique of estimating the size of large pituitary adenomas. To measure the size of a large pituitary tumor they apply a method on computed tomography (CT) scan slices of known thickness. Thus, the edge of the pituitary tumor – seen on hard copy films of the CT scan – is traced using an outlining routine on a computer and associated digitising slab. Afterwards, the measured area of the tumor is scaled and multiplied by the slice thickness in order to obtain the tumor volume for this CT slice. Finally, the overall tumor volume is obtained by calculating and summing up all volumes of the CT slices where the tumor volume is visible.

Volumetric change in pituitary adenomas over time is a critical factor in treatment decisions by physicians. Typically, the tumor volume is computed on a slice-by-slice basis using MRI patient scans obtained at regular intervals. (3D) Slicer – a free open source software platform for biomedical research – provides an alternative to this manual slice-by-slice segmentation process, which is significantly faster and less user intensive. In this study, four physicians segmented pituitary adenomas in ten patients, once using the competitive region-growing based GrowCut segmentation module of Slicer, and once purely by drawing boundaries completely manually on a slice-by-slice basis. We show and evaluate the utility of 3D Slicer in simplifying the time-consuming manual slice-by-slice segmentation while achieving a comparable accuracy. To the best of our knowledge, this is the first time the evaluation of pituitary adenoma segmentation with the free and open source medical image analysis software Slicer has been presented. Because Slicer can be downloaded and used for free, our study could be useful in clinical practice for centers different from our in which the research has been performed. Moreover, the presented GrowCut segmentation study is not limited to pituitary adenomas. GrowCut could also be used to support the segmentations of other pathologies, e.g. glioblastoma multiforme where even more time-consuming volumetry is required.

The rest of this contribution is organized as follows: *Section 2* presents the material and the methods. *Section 3* presents the results of our experiments, and *Section 4* concludes and discusses the study and outlines areas for future work.

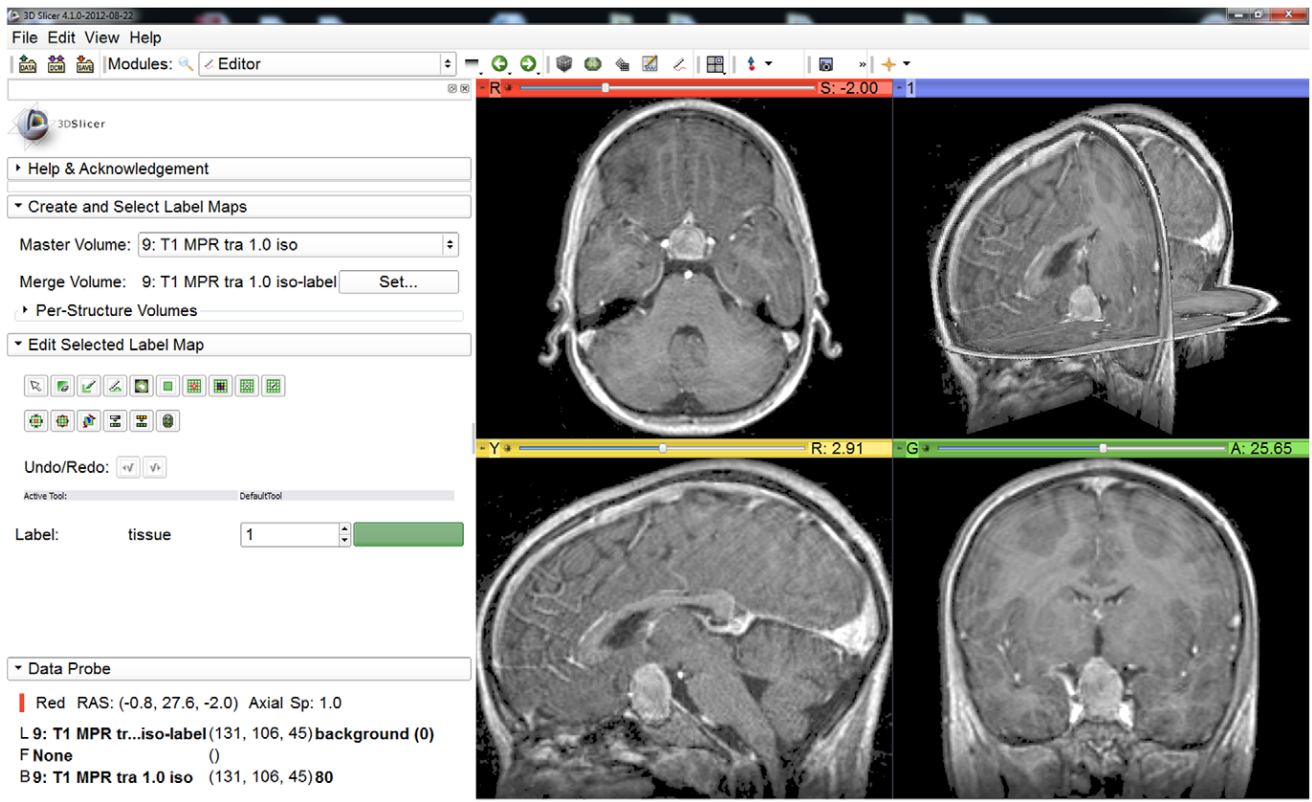

**Figure 1. 3D Slicer interface with the Slicer-Editor on the left side and a loaded pituitary adenoma data set on the right side: axial slice (upper left window), sagittal slice (lower left window), coronal slice (lower right window) and the three slices shown in a 3D visualization (upper right window).**
doi:10.1371/journal.pone.0051788.g001

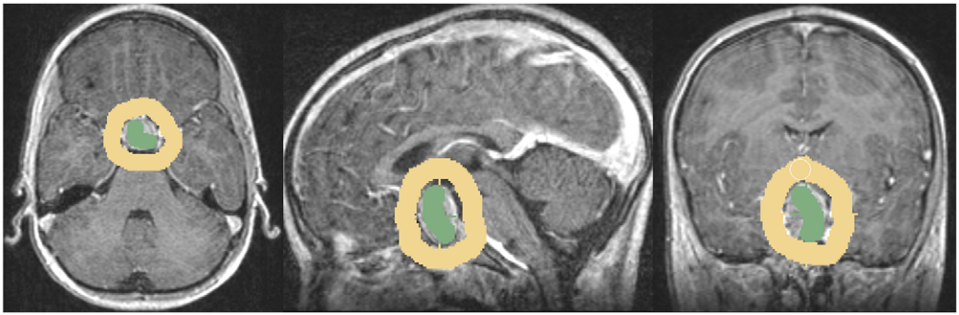

**Figure 2. These images present a typical user initialization for pituitary adenoma segmentation under Slicer with for the GrowCut algorithm: axial (left image), sagittal (middle image) and coronal (right image).** Note: the tumor has been initialized in green and the background has been initialized in yellow.
doi:10.1371/journal.pone.0051788.g002

## Materials and Methods

### Data

Ten diagnostic T1- and T2- weighted magnetic resonance imaging scans of pituitary adenomas were used for segmentation. These were acquired on a 1.5 Tesla MRI scanner (Siemens MAGNETOM Sonata, Siemens Medical Solutions, Erlangen, Germany) using a standard head coil. Scan parameters were: TR/TE 4240/84.59 msec, 3850/11 msec, 2090/4.38 msec, 690/17 msec, 480/12 msec, 479/17 msec and 450/12 msec, isotropic matrix, 1 mm; FOV, 250×250 mm; 160 sections. The segmentations have been performed mainly in coronal but for some cases also in axial slices using the homogenously contrast-enhancing structures of the pituitary adenomas in the T1/T2 scans.

### Software

The Software used in this study or the semi-automatic segmentation work was (3D) Slicer. Slicer is an open source medical image computing platform for biomedical research and freely downloadable (3D Slicer, available: http://www.slicer.org, accessed: 2012 Nov 13). To acquire the ground truth for our study, manual slice-by-slice segmentations of every data set have been performed by neurosurgeons at the *University Hospital of Marburg in Germany* (Chairman: Professor Dr. Christopher Nimsky). The physicians have several years of experience in the resection of pituitary adenomas. However, if the borders of pituitary adenomas have been very similar between consecutive slices, the physicians were allowed to skip manual segmentation for these slices. For the overall volume calculation the software interpolated the boundaries in these areas. The manual segmentation tool used for pituitary adenoma outlining provided simple contouring capabilities, and has been set-up with the medical prototyping platform MeVisLab (MeVisLab, available: http://www.mevislab.de/, accessed: 2012 Nov 13). As hardware platform we used an computer with Intel Core i5-750 CPU, 4×2.66 GHz, 8 GB RAM, with Windows XP Professional ×64 Version, Version 2003, Service Pack 2.

### GrowCut Segmentation in Slicer

GrowCut is an interactive segmentation approach that bases on the idea of cellular automaton. Using an iterative labeling procedure resembling competitive region growing, the GrowCut approach achieves reliable and reasonably fast segmentation of moderately difficult objects in 2D and in 3D. The user's initialization of GrowCut results in a set of initial seed pixels. These seed pixels in turn try to assign their labels to their pixel neighborhood which happens when the similarity measure of the two pixels weighted by the neighboring pixel's weight or "strength" exceeds its current weight. A label assignment results in an actualization of the pixel's weight as well. This labeling procedure continues iteratively until modification of the pixel labels is no longer feasible and a stable configuration has been reached. Besides the initial seed pixels – in general painted strokes on the apparent foreground and background – the GrowCut approach requires no additional inputs from the user. However, by adding additional labels in the image, the user can modify the segmentation, enabling personalization of the approach to the user. The current implementation of the GrowCut algorithm in Slicer consists of a GUI front-end to enable interactions of the user with the image and an algorithm back-end where the segmentation is computed. The GUI front-end consists of a simple to use interface which enables the user to paint directly on the image. GrowCut requires paints with at least two different colors: one for the foreground and one for the background label class. The naïve implementation of the GrowCut approach would require every single pixel to be visited within every iteration. Additionally, a pixel would need to "visit" every neighbor pixel and update the pixel's labels and strengths. Especially for large 3D images such an implementation would be computationally very expensive. Therefore, our implementation uses the following techniques for speeding up the automatic segmentation process:

- The algorithm computes the segmentation only within a small region of interest (ROI), because the user is typically interested only in segmenting out a small area in the image. This *ROI* is computed as a convex hull of all user labeled pixels with an additional margin.
- Several small regions of the image are updated simultaneously, by executing iterations involving the image in multiple threads.
- In addition, the similarity distance between the pixels are all pre-computed once and then reused.
- Moreover, the algorithm keeps track of "saturated" pixels – for "saturated" pixels the weights and therefore the labels can no longer be updated. This, on the other hand avoids the expensive neighborhood computation on those pixels. Finally, keeping track of such "saturated" pixels helps to determine when to terminate the algorithm.

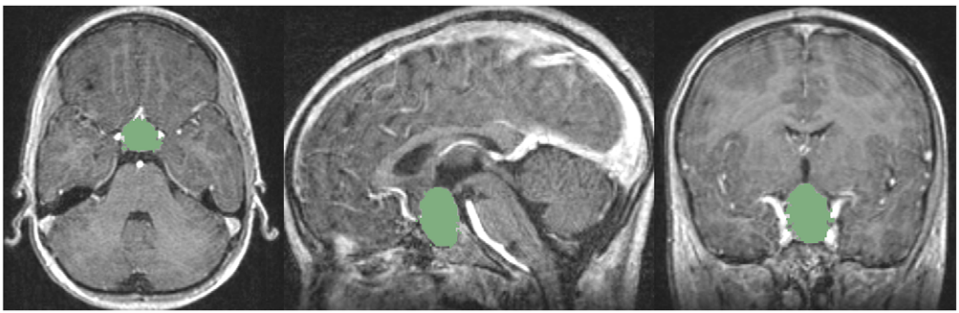

**Figure 3. These images present the segmentation result (green) of the GrowCut algorithm of Slicer: axial (left image), sagittal (middle image) and coronal (right image).** Note: the pituitary adenoma and background initialization for this segmentation result is presented in Figure 2.
doi:10.1371/journal.pone.0051788.g003

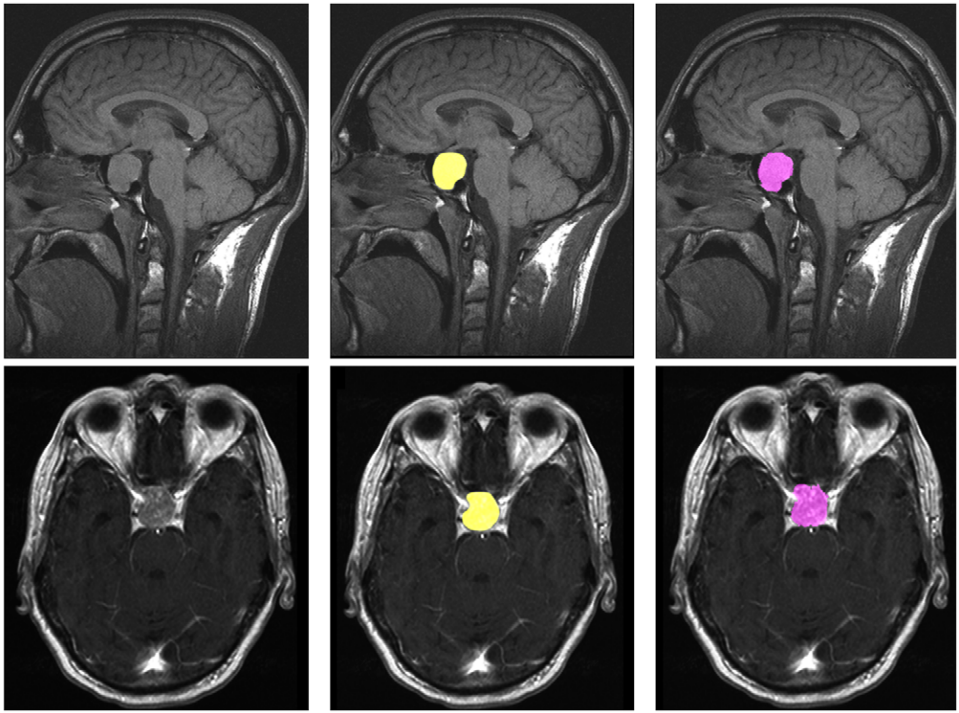

**Figure 4. These screenshots present segmentation results on a sagittal (upper row) and an axial (lower row) slice for the manual segmentation (middle images, yellow) and the Slicer-based GrowCut segmentation (right images, magenta).**
doi:10.1371/journal.pone.0051788.g004

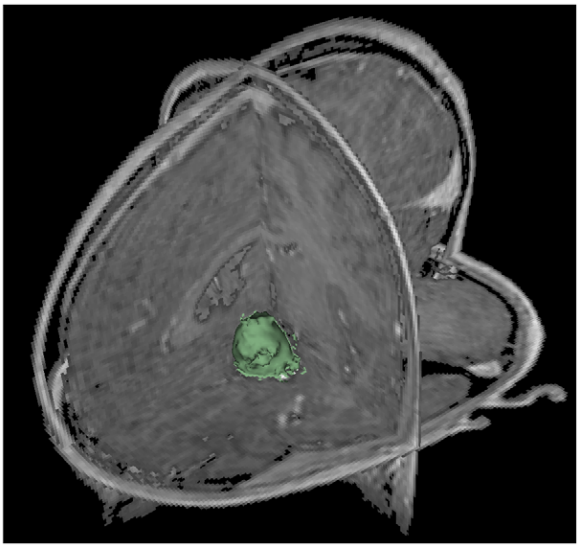

**Figure 5. This image presents the 3D segmentation result of GrowCut (green) for the tumor and background initialization of Figure 2.** After the initialization of the GrowCut algorithm under Slicer it took about three seconds to get the segmentation result on an Intel Core i7 CPU, 4×2.50 GHz, 8 GB RAM, Windows 7 Professional ×64 Version, Service Pack 1.
doi:10.1371/journal.pone.0051788.g005

## Slicer-based Pituitary Adenoma Segmentation

After testing various segmentation facilities available in Slicer, we identified that the use of GrowCut followed by additional morphological operations (like erosion, dilation, and island removal) provides the most efficient segmentation method for

**Table 1.** Direct comparison of manual slice-by-slice and Slicer-based GrowCut segmentation results for ten pituitary adenomas via the Dice Similarity Coefficient (DSC).

| Case No. | volume of pituitary adenomas (mm³) | | number of voxels | | DSC (%) |
|---|---|---|---|---|---|
| | manual | automatic | manual | automatic | |
| 1 | 6568.69 | 7195 | 72461 | 79370 | 85.87 |
| 2 | 4150.91 | 5427.76 | 4457 | 5828 | 84.36 |
| 3 | 7180.44 | 6481.12 | 35701 | 32224 | 82.11 |
| 4 | 5538.25 | 5964.5 | 61094 | 65796 | 85.1 |
| 5 | 3230.26 | 2950.45 | 22027 | 20119 | 77.51 |
| 6 | 9858.4 | 10410.8 | 67224 | 70991 | 84.46 |
| 7 | 6111.79 | 5274.89 | 52500 | 45311 | 75.6 |
| 8 | 5082.1 | 4169.32 | 56062 | 45993 | 80.1 |
| 9 | 15271.1 | 15838.9 | 104133 | 108005 | 83.41 |
| 10 | 757.007 | 1016.58 | 5162 | 6932 | 81.21 |

doi:10.1371/journal.pone.0051788.t001

pituitary adenomas for our MRI images. Therefore, the following workflow to perform pituitary adenoma segmentation has been used:

- loading the patient data set into Slicer
- initialize foreground and background for GrowCut, by drawing an area inside the pituitary adenoma and a stroke outside the tumor
- starting the automatic competing region-growing in Slicer
- after visual inspection of the results, use morphological operations like dilation, erosion, and island removal for post-editing.

The Slicer Editor module user interface, which has been used for the initialization of GrowCut is shown in Figure 1 on the left side. The right side of Figure 1 shows a pituitary adenoma after the data set is loaded into Slicer. A typical user initialization for GrowCut on the axial, sagittal and coronal cross-sections is presented in Figure 2. Finally, Figure 3 shows the results of the current Slicer GrowCut method for the initialization of Figure 2. As hardware platform for the GrowCut segmentation we used an Apple MacBook Pro with 4 Intel Core i7, 2.3 GHz, 8 GB RAM, AMD Radeon HD 6750M, Mac OS X 10.6 Snow Leopard.

### Comparison Metrics

The Dice Similarity Coefficient (DSC) [21] and [22] was used to compare the agreement between the slice-by-slice segmentations (A) and the Slicer-based segmentations (B). Therefore, we saved the segmentation results from both methods as binary volumes and calculated the relative volume overlap between the two binary volumes A and B.

### Results

The aim of this study was to evaluate the usability of Slicer for the segmentation of pituitary adenomas compared to manual slice-by-slice segmentation. Therefore, we used two metrics for an evaluation:

- The time it took for physicians to segment pituitary adenomas manually vs. using Slicer and
- the agreement between the two segmentations calculated via the Dice Similarity Coefficient.

By evaluating our results with these metrics, our assumption is that if Slicer can be used to produce pituitary adenoma segmentations that are statistically equivalent to the pure manual segmentations from physicians, and in substantially less time, then the tool is helpful for volumetric follow-ups of pituitary adenoma patients. The results of our study are presented in detail in Table 1 and Table 2, the primary conclusion of which is that Slicer-based pituitary adenoma segmentation can be performed in about two third of the time, and with acceptable DSC agreement of $81.97 \pm 3.39\%$ to slice-by-slice segmentations of physicians. Table 1 presents the segmentation results for: volume of tumor in mm³, number of voxels and Dice Similarity Coefficient for ten pituitary adenomas. Moreover, in Table 2, the summary of results: minimum, maximum, mean $\mu$ and standard deviation $\sigma$ for the ten pituitary adenomas from Table 1 are provided (note: volume is

**Table 2.** Summary of results: min, max, mean $\mu$ and standard deviation $\sigma$ for ten pituitary adenomas.

| | volume of pituitary adenomas (cm³) | | number of voxels | | DSC (%) |
|---|---|---|---|---|---|
| | manual | automatic | manual | automatic | |
| min | 0.76 | 1.02 | 4457 | 5828 | 75.60 |
| max | 15.27 | 15.84 | 104133 | 108005 | 85.87 |
| $\mu \pm \sigma$ | $6.37 \pm 3.96$ | $6.47 \pm 4.14$ | 48082.1 | 48056.9 | $81.97 \pm 3.39$ |

doi:10.1371/journal.pone.0051788.t002

presented in cm³ in Table 2). Additionally to these quantitative results, we present sample pituitary adenoma segmentation results in Figures 3, 4 and 5 for visual inspection. Figure 3 shows the results of the Slicer-based GrowCut segmentation for the tumor and background initialization of Figure 2. Figure 4 presents the direct comparison for two cases of a Slicer-based vs. the manual slice-by-slice segmentation on a sagittal (upper row) and an axial (lower row) slice. The semi-automatic Slicer-based segmentation (magenta) is shown on the right side of Figure 4 and the pure manual segmentation (yellow) is shown in the middle images of Figure 4. Finally, a 3D rendered pituitary adenoma segmentation (green) is superimposed on three orthogonal cross-sections of the data in Figure 5.

## Discussion

For accurate volumetry of cerebral pathologies like pituitary adenomas it is necessary to investigate methods that calculate the boundaries on the basis of all slices. In contrast, simpler methods – such as geometric models – provide only a rough approximation of the volume of the pathology. Especially, when accurate determination of size is of upmost importance in order to draw safe conclusions in oncology, these should not be used. Instead of all slices, geometric models use only one or several user-defined diameters, which can be achieved manually very quickly, to approximate the volume. Thereby, the volume is defined as $1/6$ $\pi d^3$ and the ellipsoid model defines the volume as $\frac{1}{6\pi\, abc}$, according to the spherical model. With $d$ as the diameter of the maximum cross-sectional area and $a$, $b$, $c$ represent the diameters in the three axes of the tumor [17]. Nobels et al. [23] measured the $x$, $y$ and $z$ radii in the frontal, sagittal and coronal plane, respectively, and assuming a spherical volume, the formula $\frac{4}{3\pi\, r^3}$ was afterwards used for the calculation of the volume – with $r$ being the mean of the $x$, $y$ and $z$ radii. Korsisaari et al. [24] estimated the size of pituitary adenoma transplants with a caliper tool from *Fred V. Fowler Co., Inc.*, by measuring the largest tumor diameter and the diameter perpendicular to this diameter (with $a$ the largest tumor diameter and $b$ the perpendicular diameter). Then, the tumor volume was calculated using $V = \frac{\pi a b^2}{6}$. Though, the clinical standard for measuring brain tumors is the Macdonald criteria [25]. These adopt uniform, rigorous response criteria similar to those in general oncology where response is defined as a ≥50% reduction of the tumor size. In general, the measure of "size" is the largest cross-sectional area (the largest cross-sectional diameter multiplied by the largest diameter perpendicular to it). Even though the semi-automatic segmentation results achieved with the GrowCut module of Slicer were reasonably good, additional editing on some slices was always required. However, these edits could be accomplished quite quickly because the GrowCut results were in close proximity of the desired pituitary adenoma boundary. Moreover, the manual segmentations by the neurosurgeons took in average about four minutes. In contrast the semi-automatic segmentation with the GrowCut implementation under Slicer took in average under three minutes, including the time needed for the post-editing of the GrowCut results.

In this study, the evaluation of pituitary adenoma segmentation with the free and open source medical image analysis software

Slicer has been presented. Slicer provides a semi-automatic, 3D segmentation algorithm called GrowCut, which is a feasible alternative to the time-consuming process of volume calculation during monitoring of a patient, for which slice-by-slice contouring has been the best demonstrated practice. In addition, Slicer offers Editing tools for a manual refinement of the results upon completion of the automatic GrowCut segmentation. Afterwards, the 3D volume of the pituitary adenomas is automatically computed and stored as an aide for the surgeon in decision making for comparison with follow-up scans. The segmentation results have been evaluated on ten pituitary adenoma data sets against manual slice-by-slice expert segmentations via the common Dice Similarity Coefficient. Summing up, the accomplished research highlights of the presented work are:

- Manual slice-by-slice segmentations of pituitary adenomas have been performed by clinical experts resulting in ground truth of tumor boundaries and estimates of rater variability.
- Physicians have been trained in segmenting pituitary adenomas with GrowCut and the Editor tools available in Slicer.
- Trained physicians segmented a pituitary adenoma evaluation set with Slicer.
- Segmentation times have been measured for the GrowCut-based segmentation under Slicer.
- The quality of the segmentations have been evaluated with the Dice Similarity Coefficient.

There are several areas of future work: For example, we plan to automate some steps of the segmentation workflow under Slicer for pituitary adenoma. For example the initialization of GrowCut could be more automated. Instead of initializing the foreground on three single 2D slices, a single 3D initialization could be used by means of generating a sphere around at the position of the user-defined seed point. In addition, the GrowCut algorithm can be enhanced with statistical information about the shape [26] and [27] and the texture [28] and [29] of pituitary adenomas to improve the automatic segmentation result. Moreover, we want to study how a Slicer-based GrowCut segmentation can be used to enhance the segmentation process of other cerebral pathologies [30], like glioblastoma multiforme. Furthermore, we are considering improving the algorithm by running the whole segmentation iteratively: After the segmentation has been performed, the result of the segmentation can be used as a new initialization for a new segmentation run and so on.

## Acknowledgments

We want to acknowledge the members of the Slicer Community and in particular Steve Pieper for their contributions, and moreover Harini Veeraraghavan and Jim Miller from GE for developing the GrowCut module for Slicer. Furthermore, the authors would like to thank the physicians Dr. med. Barbara Carl, Christoph Kappus, Dr. med. Daniela Kuhnt and Rivka Colen, M.D. for participating in this study. Finally, the authors would like to thank Fraunhofer MeVis in Bremen, Germany, for their collaboration and especially Professor Dr. Horst K. Hahn for his support.

## Author Contributions

Conceived and designed the experiments: JE. Performed the experiments: JE. Analyzed the data: JE. Contributed reagents/materials/analysis tools: JE TK CN RK. Wrote the paper: JE TK.